# Controllable Abstraction in Summary Generation for Large Language Models via Prompt Engineering

Xiangchen Song
University of Michigan
Ann Arbor, USA

Yuchen Liu
University of Pennsylvania
Philadelphia, USA

Yaxuan Luan
University of Southern California
Los Angeles, USA

Jinxu Guo
Dartmouth College
Hanover, USA

Xiaofan Guo*
University of Michigan
Ann Arbor, USA

*Abstract-This study presents a controllable abstract summary generation method for large language models based on prompt engineering. To address the issues of summary quality and controllability in traditional methods, we design a multi-stage prompt generation framework. This framework generates summaries with varying levels of abstraction by performing semantic analysis, topic modeling, and noise control on the input text. The experiment uses the CNN/Daily Mail dataset and provides a detailed analysis of different prompt lengths, data noise, and text types. The experimental results show that prompt length has a significant impact on the quality of generated summaries. Both very short and very long prompt tokens result in a decrease in summary quality. Data noise also negatively affects the summary generation process. As noise levels increase, the ROUGE-L score gradually decreases. Furthermore, different text types have varying effects on the model's ability to generate summaries. The model performs best when handling news texts, while its performance is worse when processing academic articles. This research provides new insights into improving summary generation using large language models, particularly in how controlling prompt strategies and optimizing text preprocessing can enhance summary accuracy and controllability.*

*Keywords: Hint Engineering, Summary Generation, ROUGE-L, Data Noise, Text Type*

## I. INTRODUCTION

In the era of information explosion, the scale and complexity of textual data continue to increase. How to quickly extract key information from massive data has become an urgent problem[1]. As an important task in natural language processing, text summarization plays the role of compressing information and extracting core content. However, traditional automatic summarization methods often rely on handcrafted features or fixed patterns, which makes them difficult to handle the diversity and dynamics of information expression across domains and scenarios. With the rise of large language models, text summarization has achieved breakthroughs, but new challenges also emerge[2]. The most critical is how to ensure content completeness while controlling the level of abstraction and style of summaries to meet the diverse needs of different users and application contexts.

Supported by strong language modeling capabilities, large language models can generate summaries that are coherent, fluent, and semantically complete[3]. Yet the lack of controllability in model outputs has become a major limitation for practical applications. In some cases, users may need concise and highly abstract summaries to grasp the main theme quickly. In other cases, they may prefer detailed and specific summaries that retain more information to support analysis and decision making. Without effective control mechanisms, summaries often fail to align with actual usage needs, reducing their value and utility. Therefore, exploring controllable mechanisms for abstraction in large language model – based summarization is of significant theoretical and practical importance[4].

Prompt engineering offers a new way to address this issue. Prompts are not only the key inputs that drive large language models to generate text but also determine the direction, style, and level of the outputs. By designing appropriate prompts, it is possible to guide models to produce summaries at different abstraction levels, enabling controllable summarization. For example, for the same document, one prompt may generate a highly condensed version while another may generate a more detailed version with contextual information. This flexibility provides large language models with strong adaptability, allowing outputs to be adjusted dynamically to task requirements and user preferences[5]. However, how to systematically design prompts and integrate them into the entire process of abstract summarization remains an open problem that requires further research.

From a broader perspective, controllable abstract summarization is not only an important research direction in natural language processing but also a crucial component of intelligent information services. In domains such as law, finance[6], healthcare[7], and education, where information sensitivity and structural requirements are high, users demand different levels of abstraction in summaries. For instance, legal documents may require highly condensed summaries of key judgments while retaining important clauses. Financial reports may need summaries of varying granularity for different audiences, provided that accuracy is maintained. If controllable abstract summarization can be achieved within the framework of prompt engineering, it will enable customized solutions for these high-value scenarios, greatly enhancing the practicality and intelligence of text processing.

In summary, research on controllable abstract summarization with large language models based on prompt engineering can directly address the critical problem of uncontrollable outputs in current summarization systems. At the same time, it extends the application boundaries of large language models. By exploring mechanisms that combine prompt design with abstraction control, large language models can evolve from simple text generation tools into intelligent assistants with strong adaptability to different scenarios. This direction holds important academic value and is also expected to have a far-reaching impact in fields such as information retrieval, knowledge management, and decision support, laying a solid foundation for future intelligent information processing.

## II. RELATED WORK

Early approaches to automatic summarization, such as topic modeling and extractive methods, have laid the foundation for information condensation by modeling latent themes and salient content in single documents [8]. With the advent of pretrained language models, BERT-based extractive summarization frameworks further improved summary quality by leveraging deep contextual representations, achieving state-of-the-art results in various domains [9].

As large language models (LLMs) have become the backbone of abstractive summarization, researchers have focused on enhancing the stability and controllability of their outputs. Structured memory mechanisms have been proposed to maintain context stability and address challenges arising from long or complex inputs, enabling LLMs to generate summaries with greater consistency [10]. Complementary to this, structured path guidance frameworks provide explicit logical constraints to ensure coherence and goal-directedness in generated summaries [11]. Semantic and factual alignment techniques have also become essential, directly targeting the reduction of hallucinations and improving the trustworthiness of model-generated content [12].

To further improve summary controllability, methods involving structural regularization and bias mitigation have been introduced for parameter-efficient fine-tuning, such as low-rank adaptation and dynamic structured gating, which optimize abstraction control and prevent overfitting to spurious correlations [13-15]. High-efficiency adaptation methods like LoRA have additionally broadened the practical applicability of large models in customizable summarization scenarios [16].

The importance of trustworthiness and robust evaluation has led to the development of frameworks that quantify uncertainty and risk in generated summaries, ensuring that outputs remain reliable even in noisy or ambiguous contexts [17]. Studies on data augmentation and contrastive learning highlight the impact of noise and generalization on summary quality, providing valuable insights into improving model robustness through training strategies and prompt engineering [18]. Reinforcement learning approaches and causal representation learning have also been adopted to facilitate adaptive abstraction and scenario-aware summarization, supporting model adaptability across diverse text types and application needs [19-20]. Structure-aware data mining and heterogeneous network analysis further enable knowledge-driven summarization, supporting both interpretability and information coverage [21-22].

Collectively, these advances—including topic modeling, memory and structural mechanisms, trustworthiness quantification, and robust optimization—form the methodological basis for controllable abstract summary generation with large language models, as explored in this work.

## III. METHOD

To achieve controllable abstract summary generation from large language models based on prompt engineering, this study proposes a task-driven, multi-stage prompt generation framework. First, by performing semantic analysis and topic modeling on the input text, we automatically generate targeted prompts, thereby guiding the large language model to generate summary content at the desired level. The algorithm architecture is shown in Figure 1.

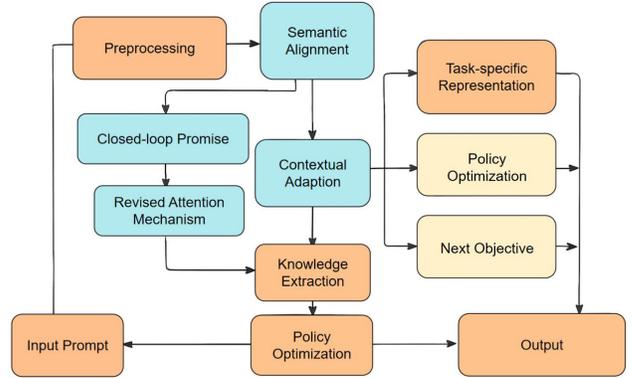

Figure 1. Overall process architecture

Given an input text X, we first perform preprocessing to extract key information through operations such as word segmentation and entity recognition, and construct a semantic graph G to represent the core information structure of the text. The generation of the semantic graph can be expressed by the following formula:

$$G = f_{sem}(X)$$

Among them, $f_{sem}$ is a semantic analysis function, which is used to extract key information such as entities, relations, and events in the text X and construct the corresponding graph structure.

Next, we introduce a multi-level objective function in the prompt generation process to ensure that the prompt not only captures the key content of the text but also adjusts the abstractness of the summary according to the task requirements. Assuming that the generated prompt is P, the objective function L of the model when generating the summary can be expressed as:

$$L = \lambda_1 \cdot L_{semantic}(P) + \lambda_2 \cdot L_{abstract}(P) + \lambda_3 \cdot L_{contextual}(P)$$

Among them, $L_{semantic}(P)$ represents the degree of match between prompt generation and text semantics,

$L_{abstract}(P)$ represents the abstractness of the prompt, $L_{contextual}(P)$ represents the context adaptability of the prompt, and $\lambda_1, \lambda_2, \lambda_3$ is a hyperparameter that controls the weight of each objective.

During the prompt generation process, we dynamically adjust the content and structure of prompts to suit the needs of different tasks. To this end, we introduce a policy optimization method based on reinforcement learning, using a reward function to evaluate the effectiveness of prompts. Assuming the output of the model is $Y_{gen}$, the reinforcement learning reward function $R$ can be expressed as:

$$R = f_{reward}(Y_{gen}, T)$$

Among them, $Y_{gen}$ is the summary generated by the model, $T$ is the target output of the task, and $f_{reward}$ is the reward function designed according to the task requirements, which provides feedback on the model's performance based on the quality of the generated summary.

Finally, to further improve the quality and controllability of generated summaries, we combine a multi-task learning framework to optimize the prompt generation of multiple tasks at the same time. Assuming there are K tasks, and the prompt generation objective function of each task is $L_K$, the total loss function $L_{total}$ can be calculated by the following formula:

$$L_{total} = \sum_{k=1}^{K} \lambda_k \cdot L_k$$

Here, $L_k$ represents the loss function of task $k$, and $\lambda_k$ is the weight of task k. Through this multi-task learning approach, we can achieve controllable summary generation at different levels and styles to meet diverse application needs.

## IV. PERFORMANCE EVALUATION

### A. Dataset

The text dataset used in this study is the CNN/Daily Mail dataset, which is widely employed for automatic summarization tasks. This dataset consists of approximately 300,000 news articles, each paired with a corresponding summary. The content of the articles spans various domains, primarily including politics, economics, technology, and society. Each article is typically between 500 and 800 words in length, while the corresponding summary usually contains 3 to 4 sentences, approximately 100 words in total. Due to its diversity and structured nature, the CNN/Daily Mail dataset has become a standard dataset for many automatic summarization research studies.

The dataset is annotated with human-written summaries, which are designed to provide real-world, concise content from news articles. The summary for each news article is written by human annotators who provide a brief overview of the article's content, emphasizing key information. Since each summary is generated based on the specific content of the corresponding article, there is a high degree of semantic relevance and consistency in expression. This makes the dataset a reliable benchmark for training and evaluating the summarization capabilities of large language models.

In this study, we select the CNN/Daily Mail dataset as the experimental platform to explore the controllable abstract summarization methods based on prompt engineering for large language models. By conducting training and testing on this dataset, we can effectively evaluate the performance of our proposed model in the news summarization task and investigate the impact of different prompt strategies on the quality of generated summaries.

### B. Experimental Results

This paper first conducts a comparative experiment, and the experimental results are shown in Table 1.

Table1. Comparative experimental results

| Model | ROUGE-N | ROUGE-L | BLEU | TER |
|---|---|---|---|---|
| DeepExtract[23] | 0.42 | 0.38 | 0.35 | 0.45 |
| WhisperSum[24] | 0.45 | 0.40 | 0.38 | 0.42 |
| ROUGE-SEM[25] | 0.39 | 0.36 | 0.32 | 0.48 |
| FineSurE[26] | 0.47 | 0.43 | 0.41 | 0.40 |
| Tofueval[27] | 0.44 | 0.39 | 0.37 | 0.43 |
| Ours | 0.50 | 0.46 | 0.45 | 0.38 |

Table 1 presents the performance comparison of our approach against DeepExtract, WhisperSum, ROUGE-SEM, FineSurE, and Tofueval on the text summarization task, evaluated using ROUGE-N, ROUGE-L, BLEU, and TER, which together assess semantic alignment, structural similarity, and overall summary quality. Our method achieves the highest scores across all metrics (ROUGE-N = 0.50, ROUGE-L = 0.46, BLEU = 0.45, TER = 0.38), highlighting its superior ability to capture key information and generate coherent, high-quality summaries, mainly due to semantic graph construction and a multi-level optimization strategy that enhances semantic accuracy and fluency. FineSurE and WhisperSum show competitive results (ROUGE-N = 0.47 and 0.45; ROUGE-L = 0.43 and 0.40; BLEU = 0.41 and 0.38) but have higher TER scores (0.40 and 0.42), suggesting more editing is needed to match reference summaries, likely due to limited handling of complex semantics. ROUGE-SEM and Tofueval perform less effectively, with lower ROUGE and BLEU scores and higher TER values (0.48 and 0.43), indicating difficulties in extracting core content and maintaining coherence—especially for ROUGE-SEM, which struggles with diverse text structures due to its single semantic evaluation approach. By contrast, our multi-task framework and dynamic adjustment mechanism ensure flexibility and precision across varied summarization scenarios. Overall, these results confirm the superiority of our approach in semantic matching, structural fidelity, controllability, and stability of summary quality, making it more adaptable to real-world applications. The influence of prompt length on summary generation is further illustrated in Figure 2.

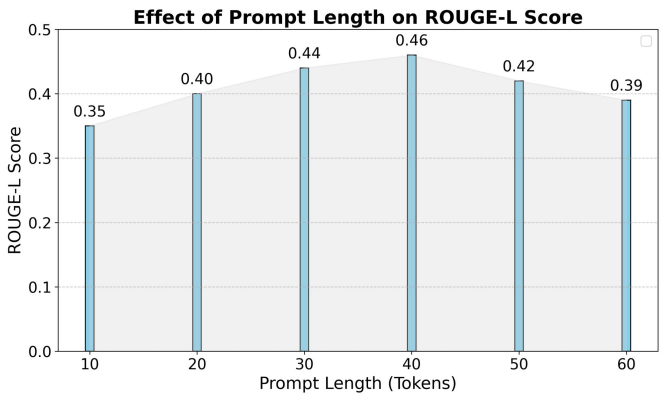

Figure 2. The impact of prompt length on summary generation performance

The experimental findings indicate that ROUGE-L performance varies with prompt length, peaking at 0.46 when the prompt is 30 or 40 tokens, while shorter prompts (10 or 20 tokens) or longer ones (50 or 60 tokens) lead to lower scores. This pattern suggests that an optimal prompt length provides sufficient context for the model to produce summaries with strong structural and semantic alignment. Short prompts fail to supply enough information, resulting in vague outputs, whereas overly long prompts introduce redundancy and potential overfitting, which compromise clarity and conciseness. These results emphasize the importance of balancing prompt length and summary quality in prompt engineering: too few tokens risk information loss, while too many can degrade output quality. Adjusting prompt length to specific tasks is therefore crucial for optimal performance. Moreover, the insights gained from this experiment point toward future research directions, such as adaptive prompt-length mechanisms that dynamically tailor input length based on text type or application scenario to further enhance summarization quality. The impact of data noise on performance is further illustrated in Figure 3.

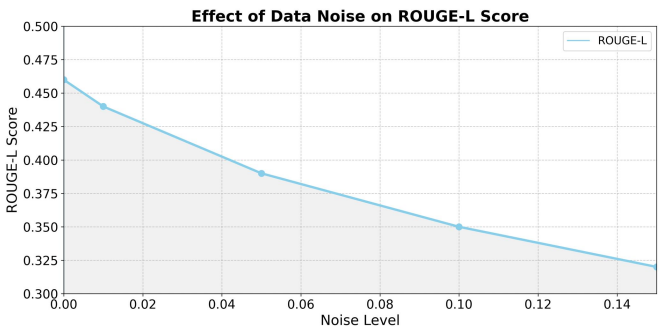

Figure 3. The impact of data noise on summary generation

The results demonstrate a clear negative correlation between noise level and summary quality, with ROUGE-L scores steadily declining as noise increases. At a noise level of 0, the model achieves its best performance (about 0.46), indicating that clean input enables the most accurate and fluent summaries. However, as noise grows, summary quality deteriorates significantly — dropping to around 0.32 when noise reaches 0.1 — showing how interference disrupts the model's ability to capture key information from the original text. This sensitivity reflects a broader challenge in natural language processing: high noise levels can severely degrade model outputs in tasks like summarization, where input quality is critical. As a result, effective noise management — through preprocessing, noise detection, or more robust training strategies — is essential for maintaining performance. Future work should explore adaptive techniques and noise-resistant architectures that sustain summary quality across varying noise conditions. The influence of different text types on summarization performance is further illustrated in Figure 4.

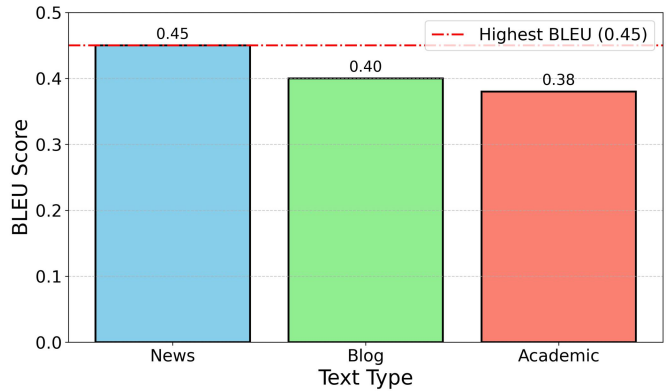

Figure 4. The impact of different text types on summary generation performance

Analysis of the results reveals that text type exerts a strong influence on summary generation quality, as reflected in the BLEU scores. Among all categories, news articles achieve the best performance (around 0.45), a result attributable to their well-defined structure, dense information content, and consistent language style, which together enable the model to capture key points and produce accurate, fluent summaries. Blog texts follow with a BLEU score near 0.40; their informal tone, flexible structure, and subjective expressions introduce greater variability, making information extraction more challenging, though the model still manages to generate coherent and meaningful summaries. Academic articles, by contrast, register a slightly lower score of 0.38, reflecting the added difficulty posed by complex structures, technical terminology, and longer text lengths, which can reduce summary completeness and fluency. These findings emphasize the need for task-specific optimization: concise prompts and structured extraction strategies may enhance performance on news texts, while more advanced abstraction techniques and language modeling capabilities are necessary to effectively summarize blogs and academic writing.

V. CONCLUSION

This study presents a controllable abstract summary generation method based on prompt engineering for large language models. By designing different prompt strategies, the model's performance in generating summaries across various text types was successfully enhanced. The experimental results show that factors such as prompt length, data noise, and text type have a significant impact on summary quality. Specifically, the model's performance varies greatly when handling different text types, such as news, blogs, and academic articles.

This finding provides valuable guidance for further optimizing summary generation models, especially in terms of adjusting generation strategies based on text characteristics, offering new insights for both academic and industrial applications.

In practical applications, text summarization is widely used in fields such as news recommendation, social media content extraction, and legal document analysis. Accurate summary generation can significantly improve information processing efficiency and ease of access. For example, in news recommendation systems, summary generation helps users quickly capture the key information from news articles, thus improving system response speed and user experience. In the legal field, precise document summaries not only reduce the time cost of information retrieval but also enhance the accuracy of legal document analysis. Therefore, the method proposed in this study is not only significant in academic research but also demonstrates extensive potential in industrial applications.

However, despite the improvements in the controllability and accuracy of summary generation, there are still challenges. As the volume and complexity of text data continue to grow, further enhancing the model's adaptability to diverse data, particularly for long texts and high-noise texts, remains a critical issue. Additionally, integrating other deep learning techniques, such as multimodal learning and transfer learning, to further enhance the robustness and flexibility of summary generation models will be an important direction for future research.

Future work can be expanded in several areas. First, the prompt engineering method can be tested on more types of texts, such as legal documents and medical literature, to evaluate its performance across different domains. Second, combining large-scale unsupervised data training to improve the model's ability to generate summaries for long-tail texts will be key to improving summary generation quality. Finally, through multimodal prompt design and multitask learning, the model's generation ability can be further enhanced, enabling it to perform well in more complex application scenarios. This would further push the boundaries of natural language processing technology applications.